\documentclass[10pt,twocolumn,letterpaper]{article}
\usepackage{iccv}
\usepackage{times}
\usepackage{amsmath}
\usepackage{amssymb}
\usepackage{booktabs}
\usepackage{caption}
\usepackage{cuted}
\usepackage{graphicx}
\usepackage{xcolor}
\usepackage{pifont}
\usepackage{float}
\usepackage[breaklinks=true,bookmarks=false,colorlinks]{hyperref}
\usepackage[capitalise]{cleveref}
\usepackage[square,numbers,sort]{natbib}
\newcommand{\cmark}{\ding{51}}%
\newcommand{\xmark}{\ding{55}}%

\newif\ifdark
\IfFileExists{/Users/vedaldi/.bash_profile}{
\immediate\write18{%
if defaults read -g AppleInterfaceStyle 2>/dev/null;
then echo \\darktrue > /tmp/displaymode.tex;
else echo \\darkfalse > /tmp/displaymode.tex; fi}
\input{/tmp/displaymode.tex}
\ifdark
\definecolor{pcolor}{HTML}{1E1E1E}
\definecolor{tcolor}{HTML}{C5C5C5}
\else
\definecolor{pcolor}{HTML}{FDF6E3}
\definecolor{tcolor}{HTML}{333333}
\fi
\pagecolor{pcolor}
\color{tcolor}
\hbadness=\maxdimen
\vbadness=\maxdimen
\vfuzz=30pt
\hfuzz=30pt
}{} 

\iccvfinalcopy
\def\iccvPaperID{9378}

\ificcvfinal\fi

\makeatletter
\renewcommand{\paragraph}{%
  \@startsection{paragraph}{4}%
  {\z@}{.5ex \@plus 1ex \@minus .2ex}{-1em}%
  {\normalfont\normalsize\bfseries}%
}
\makeatother

\makeatletter
\DeclareRobustCommand\onedot{\futurelet\@let@token\@onedot}
\def\@onedot{\ifx\@let@token.\else.\null\fi\xspace}

\def\eg{\emph{e.g}\onedot} 
\def\ie{\emph{i.e}\onedot}

\makeatother
\DeclareMathOperator{\sign}{sgn}

\begin{document}
\title{Finding an Unsupervised Image Segmenter \\ in Each of Your Deep Generative Models}

\author{
Luke Melas-Kyriazi\\
\and
Christian Rupprecht\\
\and
Iro Laina\\
\and
Andrea Vedaldi\\
\and 
University of Oxford \\
{\tt\small \{lukemk,chrisr,iro,av\}@robots.ox.ac.uk}
}

\maketitle
\ificcvfinal\thispagestyle{empty}\fi
\begin{strip}
\vspace{-5.0em}
\begin{center}
\begin{center}
\end{center}
\includegraphics[width=\linewidth]{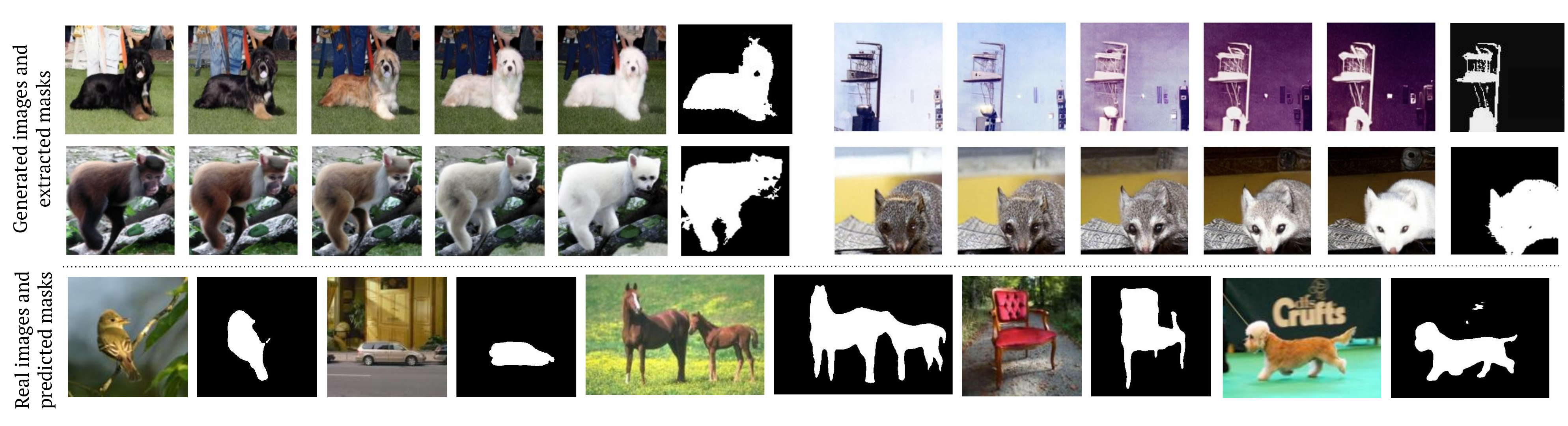}
\captionof{figure}{We propose a method to automatically find a universal latent direction in a GAN that can separate the foreground from the background. We can then generate an unlimited supply of samples with masks to train a segmentation network. The whole process is automatic and unsupervised and achieves state-of-the-art unsupervised segmentation performance.}
\label{f:splash}
\end{center}
\end{strip}

\begin{abstract}
Recent research has shown that numerous human-interpretable directions exist in the latent space of GANs. 
In this paper, we develop an automatic procedure for finding directions that lead to foreground-background image separation, and we use these directions to train an image segmentation model without human supervision.
Our method is generator-agnostic, producing strong segmentation results with a wide range of different GAN architectures.
Furthermore, by leveraging GANs pretrained on large datasets such as ImageNet, we are able to segment images from a range of domains without further training or finetuning. 
Evaluating our method on image segmentation benchmarks, we compare favorably to prior work while using neither human supervision nor access to the training data. 
Broadly, our results demonstrate that automatically extracting foreground-background structure from pretrained deep generative models can serve as a remarkably effective substitute for human supervision. 

\end{abstract}

\section{Introduction}\label{s:intro}

Self-supervised and unsupervised learning have made significant progress in the recent past, particularly with self-supervised techniques such as contrastive learning~\cite{he19momentum,chen20a-simple}.
However, most of this progress has been limited to image representations.
While useful, representations do not have an explicit meaning and are thus not immediately actionable;
instead, one still requires a supervised learning process to map the representation to a useful output, such as an image class or an object detection.

The most notable alternative to self-supervised learning is generative modelling, including variational autoencoders and generative adversarial networks.
These approaches learn to map latent codes to images, imposing simple statistical structures on the distribution of the latent codes, such as assuming an i.i.d.~Gaussian distribution.
Due to this structure, in some cases code dimensions acquire specific meanings which can be related to human-interpretable concepts (e.g., the rotation or size of an object); however, the code space in high-quality generators (e.g., BigGAN~\cite{brock_large_2018}, BigBiGANs~\cite{donahue19large}, StyleGAN~\cite{karras_style-based_2019}) is usually not easily interpretable.
Nonetheless, it is intuitive that an efficient generative process should account for the structure of natural images.
Among such structures, perhaps the most important one is the fact that images are composed of multiple objects independent of the observer.

In this paper, we thus consider the problem of learning, without supervision, a meaningful separation of images into foreground and background regions.
Our approach starts from an arbitrary, off-the-shelf high-quality generator network trained on a large corpus of (unlabeled) images.
While these generator networks are \emph{not} explicitly trained for foreground/background segmentation, we show that such a separation emerges \emph{implicitly} as a step to efficiently encode realistically-looking images.
Specifically, we design a probing scheme that can extract such foreground/background information \emph{automatically}, \ie without manual supervision, from the generator.

This scheme works as follows.
We start from a random code in latent space and learn a fixed, global offset that results in a change in the generated image. 
The offset is learned to alter the appearance of foreground and background images such that a mask can be extracted from the changes in image space. 


The resulting masks provide segmentation maps for the generated images, but they cannot yet be used to segment images from the real world.
Given a natural image, the obvious approach would be to find the corresponding code in the latent space of the generator, and then obtain a mask with our method.
Unfortunately, this inversion process is less than trivial.
In fact, recent work provides strong evidence that the expressiveness of GANs is \emph{insufficient} to encode arbitrary images~\cite{bau19seeing}, meaning that the inversion problem has no solution in general.

As we aim to build a general-purpose segmentation method, we take a different approach: we  \emph{generate} a labelled image dataset with foreground/background segmentations and use the generated dataset to train a standard segmentation network.
With this, we show that our method can successfully learn accurate foreground-background segmentation networks with no manually provided labels at all.

Compared to a recent approach by~\cite{voynov20big-gans,voynov20unsupervised}, which share some similarities to ours, we make in particular the following contributions.
First, while~\cite{voynov20big-gans,voynov20unsupervised} require manual supervision in order to extract a meaningful direction $v$ in the GAN space, in our case this direction is identified entirely automatically.
This is a key difference, because it means that our approach is unsupervised.
Second,~\cite{voynov20big-gans} focus on a particular generator network, BigBiGAN~\cite{donahue19large}, whereas we consider a large family of different networks.
The importance of doing so is that it shows that the method can discover meaningful code space directions for multiple models with no need for model-specific manual intervention.
In this manner, we show that learning to generate images is conductive to learning foreground/background segmentation, even for generic generator models that are not trained specifically to achieve such an effect.

We also show that our method can learn a `universal' foreground-background segmenter.
By this, we mean that, by constructing our image segmenter from a generator network trained on a generic large-scale dataset such as ImageNet, applied to segmenting objects and salient regions in specialized datasets such as CUB200~\cite{WelinderEtal2010} and the Oxford Flowers Dataset~\cite{Nilsback09a}, still attains very strong foreground-background segmentation results despite the obvious statistical misalignment between the training and testing data. 
Furthermore, when evaluated on saliency detection benchmarks, our method approaches (and sometimes even exceeds) the performance of supervised and handcrafted saliency detection methods using neither supervision nor access to the training data.

Finally, we find that the segmentation performance directly correlates with the quality of the underlying GAN, which means that foreground/background separation is an important concept in learning generative models.
Additionally, segmentation performance could be used as an objective metric to assess a generative model --- which is still difficult and currently often relies on human experiments.
\section{Related work}\label{s:related}




Our method is related to generative models, object segmentation, and saliency detection, as discussed next.

\paragraph{Interpreting Deep Generative Models.}

Several works have proposed methods for decomposing the latent space of a generative model into interpretable or disentangled directions.
Early work included Beta-VAE~\cite{higgins2017beta}, which modified the variational ELBO in the original VAE formulation, and InfoGAN~\cite{chen2016infogan}, which maximized the mutual information between a subset of the latent code and the generated data.
Later work has sought to disentangle factors of variation by mixing latent codes~\cite{hu2018disentangling}, adding additional adversarial losses~\cite{mathieu2016disentangling}, and using contrastive learning~\cite{ren2021generative}. 

Our work follows a recent line of research that looks for structure in large, pretrained generative models.~\cite{shen2020closed} performs a direct decomposition of model weights to find disentangled directions, while~\cite{peebles2020hessian} penalizes nonzero second-order interactions between different latent dimensions, and~\cite{voynov20unsupervised} finds interpretable directions by introducing an additional reconstruction network. 

Differently from the works above, we conduct a deep study of \textit{one} specific type of structure (foreground/background separation) encoded in the latent space.
Other works have taken this approach in the context of extracting 3D structure from 2D images~\cite{nguyen2019hologan}; Inverse Graphics GAN~\cite{lunz2020inverse} uses a neural renderer to recover 3D (voxel-based) representations of scenes, and GAN2Shape~\cite{pan20202d} exploits viewpoint and lighting variations in generated images to recover 3D shapes.

\paragraph{Unsupervised Object Segmentation.}

Prior work on unsupervised object segmentation can be divided into two categories: those that employ generative models to obtain segmentation masks and those that employ purely discriminative methods such as contrastive learning~\cite{Ji_2019_ICCV,ouali_autoregressive_2020}.
Here, we focus on generative approaches. 

Nearly all generative approaches are based on the idea of decomposing the generative process in a layer-wise fashion; in general, the foreground and background of an image are generated separately and then combined to obtain a final image.
Specifically, ReDo~\cite{chen_unsupervised_2019} trains a generator to re-draw new objects on top of old objects, and enforces realism through adversarial training.
{}\cite{bielski2019emergence} generates a background, a foreground, and a foreground mask separately and composite them together; they prevent degenerate outputs (\ie the foreground and background being the same) by randomly shifting the foreground relative to the background.
Copy-Paste GAN~\cite{arandjelovic2019object} receives two images as input and copies parts of one image onto the other.
OneGAN~\cite{benny_onegan_2020} learns to simultaneously generate, cluster, and segment images with a combination of GANs, VAEs, and additional encoders. 

Our approach is based on generative modeling, but it differs from most other approaches in that we seek to find foreground/background structure \textit{implicitly} encoded in (standard, non-layer-wise) GANs rather than encoding it \textit{explicitly}.
This enables us to leverage pretrained generators with highly-optimized architectures trained on millions of high-resolution images, rather than developing a new GAN architecture for this specific task.

Furthermore, approaches based on explicit image decomposition assume that the foreground and background of an image are independent.
This assumption is clearly violated in real-world data (e.g., birds are more likely to appear with blue backgrounds), meaning that these methods are fundamentally limited. Our approach does not rely on such an independence assumption.

One very recent work that shares these advantages is~\cite{voynov20big-gans}, which employs a pretrained BigBiGAN~\cite{donahue19large} generator.
{}\cite{voynov20big-gans} uses the method from~\cite{voynov20unsupervised} to decompose the latent space into interpretable directions, manually handpicks a direction that separates the foreground and background, and then uses the direction to train a segmentation model.
Although this method does not require pixel-level labels, it is still supervised in the sense that a person must manually select the desired latent direction.
Further, it is not clear that such a procedure (that is, the method from~\cite{voynov20unsupervised}) will find an effective foreground/background-separating direction for other GANs.
Differently from their method, ours is completely unsupervised, applies to arbitrary GANs, and delivers superior performance across object segmentation and saliency detection benchmarks. 

\paragraph{Saliency Detection.}

Object segmentation is closely related to saliency detection, the problem of finding significant (salient) objects in an image. 
The past few years have seen some research into unsupervised/weakly-supervised saliency detection~\cite{zhang_deep_2018,nguyen_deepusps_2019,zeng_multi-source_2019}.
These methods work by ensembling strong hand-crafted priors and distilling them into a deep network.
In practice, they also initialize their networks with pretrained (supervised) image classifiers or semantic segmentation networks. 


Finally, our method can be viewed from the perspective of learning from synthetic data.
For example, one widely-studied line of research~\cite{hoffman18cycada:,Tsai_2018_CVPR,Zou_2018_ECCV,Vu_2019_CVPR,toldo2020unsupervised} tackles the task of semantic segmentation by training on data generated from video games (\eg GTA5).
With regard to adversarially-generated training data specifically~\cite{shrivastava2017learning} uses a GAN-like network to enhance the realism of synthetic images while preserving label information.

Although we train our segmentation network using generated images only, we show in the experiments below that it generalizes to real-world images without the need for additional adaptation.

\begin{figure}[t]
\centering
\includegraphics[width=0.47\textwidth]{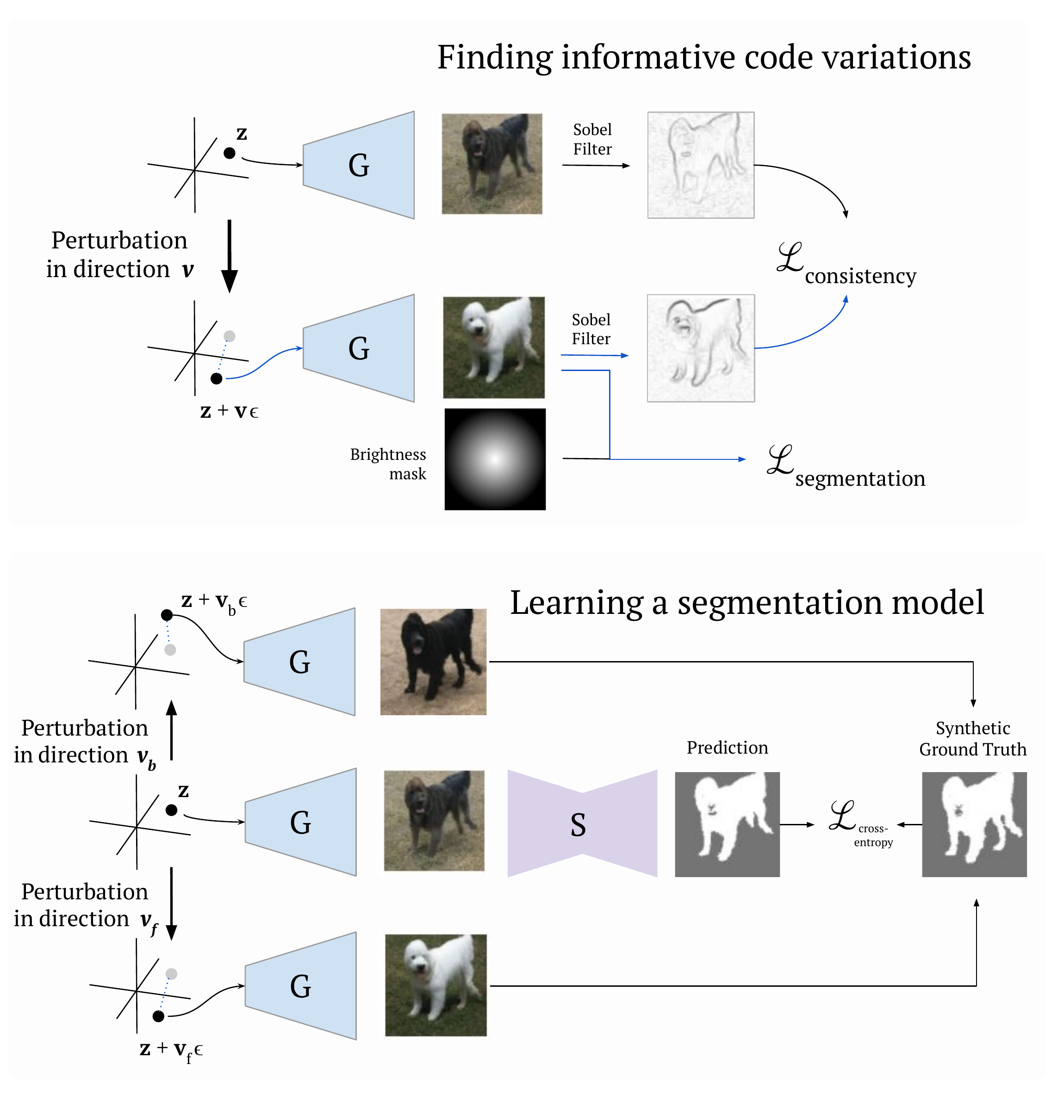}
\caption{Our unsupervised segmentation pipeline, consisting of two stages. First (above), a direction is identified in the latent space of a deep generative model ($G$) that separates the foreground and background of generated images by changing their relative brightness. Second (below), a synthetic dataset is generated using this direction (or two of these directions) and is used to train a separate segmentation network ($S$). This network can then be applied to unseen real-world data without further training.}%
\label{fig:diagram}
\end{figure}

\section{Method}\label{s:method}

Let $x\in\mathbb{R}^{3\times H \times W}$ be a (color) image.
A \emph{generator} (network) is a function $G : \mathbb{R}^D \rightarrow \mathbb{R}^{3\times H\times W}$ that maps code variables $z$ to images $x=G(z)$.
Optionally, some generative models come with an encoder function $E: \mathbb{R}^{3\times H\times W} \rightarrow \mathbb{R}^{D}$ which computes an approximate inverse of the generator (i.e.~$G(E(x))\approx x$).

A challenge in generating images is that individual pixels exhibit complex correlations, caused by the fact that the images are obtained as the composition of a number of different objects.
For example, all pixels that belong to a human face have a similar color, characteristic of one's complexion.
However, the correlation is much less strong between pixels that belong to \emph{different} objects.
This is because, while object in a scene are not entirely independent, their correlation is much weaker than within the structure of objects.

Intuitively, an image generator must learn to account for such correlations in order to generate realistically-looking images.
In particular, we expect the generator to somehow capture the idea that pixels that belong to the same object have a related appearance, whereas the appearance of pixels that belong to different objects or, as it may be, to a foreground object and its background, should be much more statistically independent.

Given a generator function $G$, it is then natural to ask whether such correlations can be extracted and used not just for the purpose of generating images, but also for analyzing them.
In order to explore this idea, we consider perturbing the code $z$ via a small increment $\epsilon v \in\mathbb{R}^D$, where $\epsilon\in\mathbb{R}$ and $v\in\mathbb{S}^{D-1}$ is a unit vector.
Because the dimension $D$ of the embedding space is typically much smaller than the dimension $3HW$ of the generated images, codes provide highly-compressed views of the data (for example, $D=120$ for BigBiGAN~\cite{donahue19large} and the self-conditioned GAN).
As such, most changes in the code are likely to affect most if not all pixels in the image.
However, if the generator does learn to compose objects, then it should be possible to find a variation $v$ that only affects a portion of the image, corresponding to an individual object.
Intuitively, if we can find such a selective variation, we may use it to highlight and segment an object in the image.

Empirically, we find that the situation is not as simple.
Specifically, it is not easy to find changes in the code that leave part of the pixels exactly constant while changing other pixels.
However, inspired by~\cite{voynov20unsupervised}, we find that there are directions that affect foreground and background regions in a systematic and characteristic manner.
Furthermore, we show that these directions are `universal', in the sense that the \emph{same} $v$ works for \emph{all} codes $z$, and are thus characteristic of a given generator network $G$.

\subsection{Finding informative code variations}

Next, we introduce an automated criterion to select informative changes $v$ in code space.
To this end, we consider an image $x=G(z)$ generated from a random code $z\sim\mathcal{Z}$, where $\mathcal{Z}$ is the code distribution characteristic of the generator (\eg an i.i.d.~Gaussian).
We then consider a modified image $x'=G(z+\epsilon v)$ and observe the change $x\rightarrow x'$.

We compare the two images using two criteria.
The first one preserves the \emph{structure} of the image $x$.
We capture the latter by imposing that $x$ and $x'$ generate approximately the same edges when fed to a simple edge detector.
The intuition is that we wish $v$ to affect the appearance of objects without changing their shape.
By preventing objects from `moving around' the image or deforming, we make it significantly easier to extract an image segmentation from the change $x\rightarrow x'$.
This loss takes the form:
$$
\mathcal{L}_s(v)
=
\frac{1}{N} \sum_{i=1}^N
\|
S(G(z_i+\epsilon v)) - S(G(z_i))
\|^2
$$
where $z_i\sim\mathcal{Z}$ and $S$ is the Sobel-Feldman operator:
$$
[S(x)]_{ij} =
\sum_{c=1}^3
(g \ast x_{c::})_{ij}^2 +
(g^\top \ast x_{c::})_{ij}^2,
$$
and
$
g
=
\begin{bmatrix}
1 & 2 & 1
\end{bmatrix}
\cdot
\begin{bmatrix}
1 & 0 & -1
\end{bmatrix}^\top.
$

This loss encourages $x$ and $x'$ to be similar. 
We thus also need a loss that encourages the direction $v$ to explore a non-zero change of the image.
Inspired by~\cite{voynov20big-gans}, we consider an image contrast variation and additionally exploit the photographer bias, that objects are often placed in the middle of the image.
This is captured by the loss:
$$
\mathcal{L}_c(v)
=
\frac{1}{N} \sum_{i=1}^N
\sum_{c=1}^3 \langle G(z+\epsilon v), r \rangle
$$
where $r\in\mathbb{R}^{H\times W}$ is a `radial' prior:
$$
    r_{ij}
    =
    1 - \frac{1}{\alpha}
    \left[
    \left(i - \frac{H+1}{2}\right)^2 +
    \left(j - \frac{W+1}{2}\right)^2
    \right]
$$
with normalization factor $\alpha=\frac{1}{4}\sqrt{(H-1)^2+(W-1)^2}$ that linearly interpolates from $1$ in the center of the image to $-1$ at the boundary.
This encourages finding a direction $v$ that changes the brightness in the center of the image opposite to the border.


In order to learn $v$, the two losses are combined
\begin{equation} \label{eq:opt}
    \mathcal{L}(v)=
    \mathcal{L}_c(v) + \lambda \mathcal{L}_s(v)
\end{equation}
where $\lambda$ is a weighting factor.

We note that, given this fully-automatic procedure, the latent code direction $v$ may be thought of as a function of the generator $G$ and the weighting factor $\lambda$.

\subsubsection{Combining informative codes}

Optimizing \cref{eq:opt} with $\lambda > 0$ encourages the network to produce a shift $v$ that brightens the foreground and darkens the background of an image. However, there is no constraint that $\lambda$ need be positive; by negating $\lambda$ and optimizing a second time, we obtain another direction $v$ that shifts the foreground dark and the background light.

Although using only one direction suffices for our method, we find that we can improve performance by using both. As a result, for the remainder of the paper, let $v_l$ represent the direction that shifts the foreground lighter, and $v_d$ represent the direction that shifts the foreground darker.

\subsection{Learning a segmentation model}

Once the latent directions $v_d$ and $v_l$ have been found, the process of extracting a segmentation mask is straightforward: we label as foreground regions the pixels in which the image generated with the foreground-lighter shifted latent code is lighter than the image generated with the foreground-darker shifted latent code.
That is, for a generated image $x = G(z)$, we have:
\begin{equation} \label{eq:mask_gen_dual}
M(z) = \sign(G(z + \epsilon v_l) - G(z + \epsilon v_b))
\end{equation}
Alternatively, if we use only a single direction $v_l$ or $v_b$, $M(z)$ is set to either:
\begin{equation}\label{eq:mask_gen_single}
G(z + \epsilon v_l) - G(z) ~~~ \text{or} ~~~ G(z) - G(z + \epsilon v_b)
\end{equation}
Given the learned direction $v$, we use it to generate a training set as follows:
$$
\mathcal{D} = \{
    (G(z_i), M(z_i)):~ z_i \sim\mathcal{Z}, ~i = 1, \ldots 
\}.
$$
This dataset may then be used to train any dense segmentation network $\Psi$ (i.e., a UNet~\cite{ronneberger15u-net:}) in the standard fashion.
That is, we minimize the pixel-wise binary cross-entropy loss between the segmentation output $\Psi(G(z)) \in \mathbb{R}^{H\times W}$ of the network and the (synthesized) mask $M(z)$:
$$
\mathcal{L}(\Psi|z) = 
- \frac{1}{HW} \sum_{u \in [H] \times [W]} \log p(\sign(M_u(z))|\Psi_u(G(z)))
$$
where $p(m|s) = m \sigma(s) + (1-m)\sigma(-s)$, $u$ is a pixel index and $\sign$ is the sign function.
Unlike previous object segmentation methods, our method requires no additional losses or constraints to ensure the stability of training.
By sampling $z$, we can generate an `infinite' dataset for learning the network $\Psi$:
$$
\mathcal{L}(\Psi)
 = E_{z\sim\mathcal{Z}} [ \mathcal{L}(\Psi|z) ]
$$

\textit{Note:} Although we described the procedure above for unconditional GANs because their use makes our method completely unsupervised, our method applies just as well for weakly-supervised conditional GANs, where the generator $G(z,y)$ also depends on a class label.
In this case, we can apply the exact same procedure for optimization and training, picking the labels $y$ uniformly at random.

\subsubsection{Refining the generated dataset}

An advantage of training with GAN-generated data is that the dataset size is infinite, which means that one is free to curate one's dataset and discard uninformative training examples. In our case, similarly to \cite{voynov20big-gans}, we found that it was helpful to refine the dataset by (1) discarding images with masks that were too large, (2) discarding images for which the latent code shift did not produce a large change in brightness, and (3) removing small connected components from the mask. For purposes of comparison, we kept the refinement process exactly the same as that \cite{voynov20big-gans}. The exact details are given in the Supplementary Material.

\begin{table}[ht!]
  \small
  \centering
  \setlength{\tabcolsep}{4.5pt}
  \begin{tabular}{@{\hspace{-3pt}}c@{\hspace{2pt}}l|ccc|ccc}
  \toprule
  & \multicolumn{1}{c}{} & \multicolumn{3}{c}{\textbf{DUTS}} & \multicolumn{3}{c}{\textbf{ECSSD}} \\ 
  & & Acc  & IoU  & $F_\beta$  & Acc  & IoU  & $F_\beta$          \\ \midrule
  \multicolumn{8}{c}{\textit{\footnotesize Supervised Methods}} \\ \midrule
  \cite{huo2019short}         & Hou et al.        & 0.924 & -   & 0.729 & 0.930 & -   & 0.880 \\
  \cite{luo2017nonlocaldeep}  & Luo et al.        & 0.920 & -   & 0.736 & 0.934 & -   & 0.891 \\
  \cite{zhang2017aggregating} & Zhang et al.      & 0.902 & -   & 0.693 & 0.939 & -   & 0.883 \\
  \cite{zhang2017uncertain}   & Zhang et al.      & 0.868 & -   & 0.660 & 0.920 & -   & 0.852 \\
  \cite{wang2016stagewise}    & Wang et al.       & 0.915 & -   & 0.672 & 0.908 & -   & 0.826 \\
  \cite{li2016deepsaliency}   & Li et al.         & 0.924 & -   & 0.605 & 0.840 & -   & 0.759 \\
  \midrule
  \multicolumn{8}{c}{\textit{\footnotesize Handcrafted Methods}} \\ \midrule
  \cite{zhu2014rbd}  & RBD        & 0.799  & -    & 0.510  & 0.817 & -      & 0.652       \\
  \cite{li2013dsr}   & DSR        & 0.863  & -    & 0.558  & 0.826 & -      & 0.639       \\
  \cite{jiang2013mc} & MC         & 0.814  & -    & 0.529  & 0.796 & -      & 0.611       \\
  \cite{zhou2015hf}  & HS         & 0.773  & -    & 0.521  & 0.772 & -      & 0.623       \\\midrule
  \multicolumn{8}{c}{\textit{\footnotesize Deep Ensembles of Handcrafted Methods}} \\ \midrule
  \cite{zhang_2017_sbf} & SBF             & 0.865  & -    & 0.583  & 0.915 & -      & 0.787       \\
  \cite{zhang_deep_2018} & USD$^{**}$      & 0.914  & -    & 0.716  & 0.930 & -      & 0.878       \\
  \cite{nguyen_deepusps_2019} & USPS$^{**}$$^{\dagger}$ & 0.938  & -    & 0.736  & 0.937 & -      & 0.874       \\ \midrule
  \multicolumn{8}{c}{\textit{\footnotesize Weakly-Supervised Methods}} \\ \midrule
  \cite{voynov20big-gans} &  Voynov$^*$             &  0.878          &  0.498          &  -                              &  0.899             & 0.672          &  -      \\
  \cite{voynov20big-gans} &  Voynov$^*$$^{\diamond}$ &  0.881          &  0.508          &  0.600                          &  0.906             &  0.685         &  0.790  \\ \midrule
  \multicolumn{8}{c}{\textit{\footnotesize Unsupervised Methods}} \\ \midrule
  &  Ours                   &  0.893 &  0.528 &  0.614                 &  0.915    & 0.713 &  0.806       \\ 
  \bottomrule
\end{tabular}

\vspace{1mm}

\setlength{\tabcolsep}{3.5pt}
\begin{tabular}{@{\hspace{-3pt}}c@{\hspace{2pt}}l|ccc|ccc} 
  \toprule
  & \multicolumn{1}{c}{} & \multicolumn{3}{c}{\textbf{CUB}} & \multicolumn{3}{c}{\textbf{Flowers}} \\ 
  &                         &  Acc            &  IoU            &  $\text{max}F_\beta$& Acc &  IoU                                                                        &  $\text{max}F_\beta$    \\  
  \midrule
  \multicolumn{8}{c}{\textit{\footnotesize Weakly-Supervised Methods}} \\ \midrule
  \cite{voynov20big-gans}           &  Voynov$^*$             &  0.930          &  0.683          &  0.794& 0.765              &  0.540                                                                      &  0.760                                \\ 
  \cite{voynov20big-gans}           &  Voynov$^*$$^{\diamond}$ &  0.931 &  0.693 &  0.807            &  0.777                                                                      &  0.529                                                     &  0.672                                 \\ 
  \midrule
  \multicolumn{8}{c}{\textit{\footnotesize Unsupervised Methods}} \\ \midrule
  \cite{bielski_emergence_2019}     &  PertGAN             &  -              &  0.380          &  -                         &  -                                                                          &  -                                                         &  -                                    \\ 
  \cite{chen_unsupervised_2019}     &  ReDO                   &  0.845          &  0.426          &  -                         &  0.879& 0.764& -                                    \\ 
  \cite{xia2017wnet}                &  WNet$^{\dagger}$                   &  -              &  0.248          &  -                         &  -                                                                          &  -                                                         &  -                                    \\ 
  \cite{kanezaki_unsupervised_2018} &  UISB                   &  -              &  0.442          &  -                         &  -                                                                          &  -                                                         &  -                                    \\ 
  \cite{ji19invariant}              &  IIC-seg                &  -              &  0.365          &  -                         &  -                                                                          &  -                                                         &  -                                    \\ 
  \cite{benny_onegan_2020}          &  OneGAN                 &  -              &  0.555          &  -                         &  -                                                                          &  -                                                         &  -                                    \\ 
  \midrule
  &  Ours   \hspace{8mm}                &  0.921 &  0.664 &  0.783                     &  0.796                                                             &  0.541                                            &  0.723                                 \\ 
  \bottomrule
\end{tabular}

\vspace{1mm}

\setlength{\tabcolsep}{6pt}
\begin{tabular}{@{\hspace{-3pt}}c@{\hspace{2pt}}l|ccc} %
\toprule
                        & \multicolumn{1}{c}{}    & \multicolumn{3}{c}{\textbf{DUT-OMRON}} \\ 
                        &                         &  Acc            &  IoU            &  $\text{max}F_\beta$ \\ \midrule
\cite{voynov20big-gans} &  Voynov$^*$             &  0.856          &  0.453                          &  -             \\ 
\cite{voynov20big-gans} &  Voynov$^*$$^{\diamond}$ &  0.859          &  0.460                          &  0.533         \\ \midrule
                        &  Ours                   &  0.883 &  0.509                 & 0.583 \\
\bottomrule
\end{tabular}

\caption{\small Performance on three saliency detection benchmarks (DUTS, ECSSD, DUT-OMRON) and two object segmentation benchmarks (CUB, Flowers). 
$^{**}$ initializes with a pretrained (supervised) network. 
$^{\dagger}$ CRF post-processing. $^{\diamond}$ our implementation.
}
\label{table:benchmark}
\end{table}
\begin{table*}[ht]
  \small
  \centering
  \begin{tabular}{cl|cc|cc|cc|cc|cc|cc}
          \toprule
      &      \multicolumn{1}{c}{}  & & \multicolumn{1}{c}{} & \multicolumn{2}{c}{\textbf{CUB}} & \multicolumn{2}{c}{\textbf{Flowers}} & \multicolumn{2}{c}{\textbf{DUT-OMRON}} & \multicolumn{2}{c}{\textbf{DUTS}} & \multicolumn{2}{c}{\textbf{ECSSD}} \\ 
  & \multicolumn{1}{c}{}                               &  Dataset      &  Res. &  Acc            &  IoU            &  Acc            &  IoU            &  Acc            &  IoU            &  Acc            &  IoU            &  Acc            &  IoU         \\ \midrule
  \cite{pmlr-v70-odena17a} & \textit{ACGAN}            &  TinyImageNet &  64                       &  0.682          &  0.265          &  0.572          &  0.266          &  0.642          &  0.190          &  0.647          &  0.191          &  0.652          &  0.276  \\
  \cite{brock_large_2018} & \textit{BigGAN}            &  TinyImageNet &  64                       &  0.853          &  0.257          &  0.723          &  0.284          &  0.844          &  0.213          &  0.842          &  0.224          &  0.811          &  0.332  \\
  \cite{lim2017geometric} & \textit{GGAN}              &  TinyImageNet &  64                       &  0.818          &  0.366          &  0.697          &  0.315          &  0.782          &  0.221          &  0.783          &  0.235          &  0.766          &  0.316  \\
  \cite{zhang2019sagan} & \textit{SAGAN}               &  TinyImageNet &  64                       &  0.828          &  0.376          &  0.732          &  0.351          &  0.808          &  0.235          &  0.806          &  0.246          &  0.788          &  0.327  \\
  \cite{zhang2019sngan} & \textit{SNGAN}               &  TinyImageNet &  64                       &  0.849          &  0.357          &  0.751          &  0.374          &  0.816          &  0.216          &  0.814          &  0.217          &  0.795          &  0.292  \\
  \cite{zhang2019sagan} & \textit{SAGAN}               &  ImageNet     &  128                      &  0.871          &  0.336          &  0.608          &  0.085          &  0.856          &  0.250          &  0.860          &  0.282          &  0.814          &  0.340  \\
  \cite{zhang2019sngan} & \textit{SNGAN}               &  ImageNet     &  128                      &  0.881          &  0.378          &  0.703          &  0.304          &  0.860          &  0.305          &  0.854          &  0.300          &  0.837          &  0.432  \\
  \cite{Kang2020contragan} & \textit{ContraGAN}        &  ImageNet     &  128                      &  0.857          &  0.159          &  0.661          &  0.088          &  0.858          &  0.075          &  0.870          &  0.149          &  0.805          &  0.204  \\
  \cite{liu2020selfconditioned} & \textit{UnCondGAN}   &  ImageNet     &  128                      &  0.734          &  0.217          &  0.494          &  0.049          &  0.698          &  0.127          &  0.729          &  0.158          &  0.681          &  0.198  \\
  \cite{liu2020selfconditioned} & \textit{SelfCondGAN} &  ImageNet     &  128                      &  0.869          &  0.459          &  0.670          &  0.238          &  0.800          &  0.280          &  0.806          &  0.297          &  0.806          &  0.412  \\
  \cite{brock_large_2018} & \textit{BigGAN}            &  ImageNet     &  128                      &  0.886          &  0.367          &  0.731          &  0.318          &  0.883          &  0.316          &  0.876          &  0.303          &  0.848          &  0.424  \\
  \cite{donahue19large} & \textit{BigBiGAN}            &  ImageNet     &  128                      &  \textbf{0.912} &  \textbf{0.601} &  \textbf{0.773} &  \textbf{0.479} &  \textbf{0.878} &  \textbf{0.451} &  \textbf{0.890} &  \textbf{0.486} &  \textbf{0.905} &  \textbf{0.663}  \\
  \bottomrule
  \end{tabular}
  \caption{A comparison of segmentation model performance across a wide range of generator architectures, using a foreground-lighter shift ($v_l$). All hyperparameters are kept constant across generators.}
  \label{table:gan_comparison}
  \end{table*}
\begin{figure}
  \centering
  \includegraphics[width=0.47\textwidth]{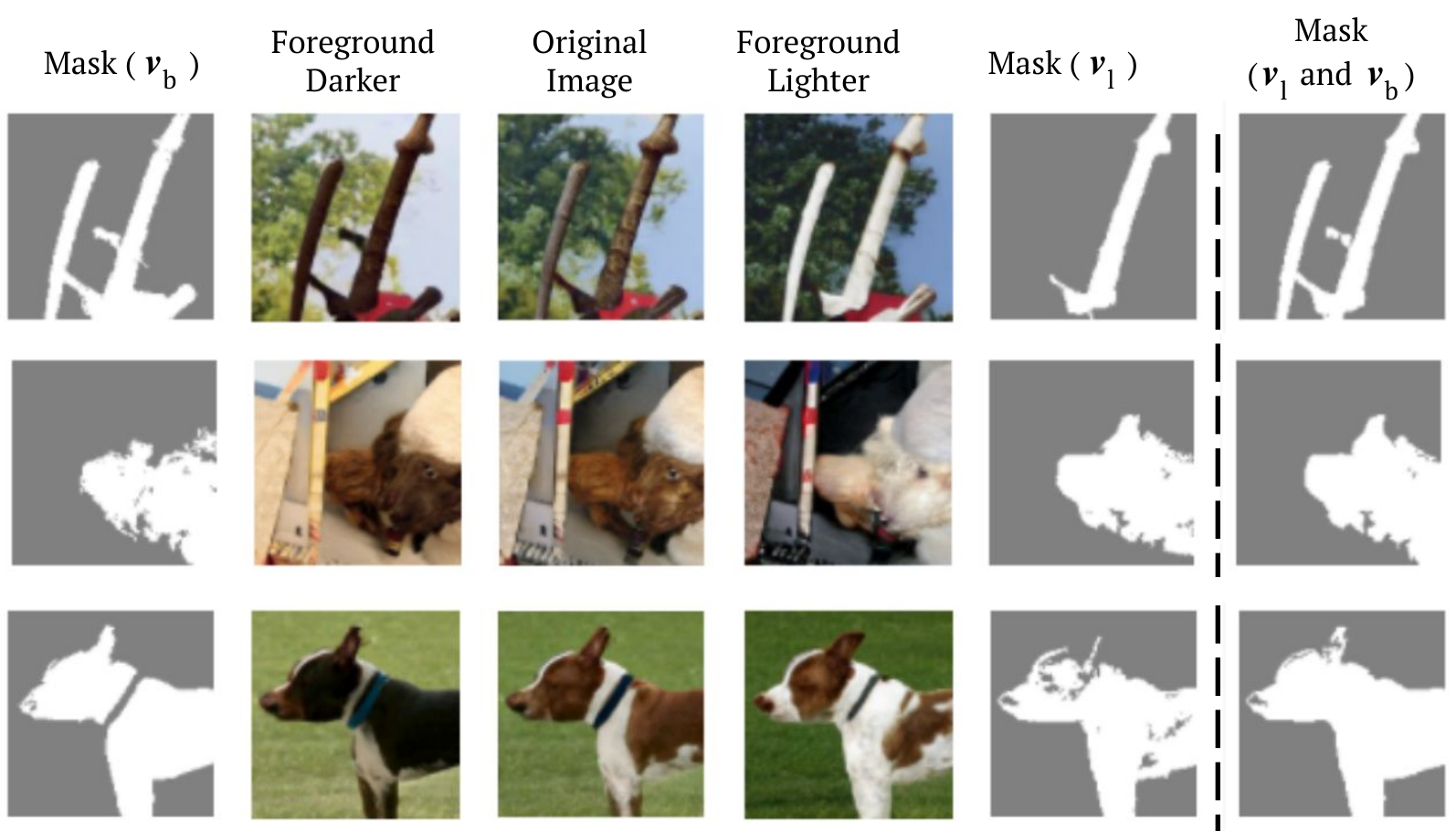}
  \caption{A comparison of perturbed images and their corresponding masks for $v_b$, which shifts the foreground darker, $v_l$, which shifts the foreground lighter, and the combination $v_b$ and $v_l$. Using the combination of both directions yields visually superior segmentations.}\label{fig:direction_comparison}
\end{figure}

\begin{table}[]
\small 
\centering
\begin{tabular}{cccccc} \toprule
Dataset   &  & $v_l$ & $v_d$ & $v_l$ \& $v_d$ & ensemb.    \\ \midrule
CUB       & \textit{Acc}                  & 0.912                          & 0.912                          & \textbf{0.921}     & \textbf{0.921} \\
          & \textit{IoU}                  & 0.601                          & 0.631                          & \textbf{0.664}     & 0.650                           \\
          \midrule
Flowers   & \textit{Acc}                  & 0.773                          & 0.806                          & 0.796                               & \textbf{0.799} \\
                           & \textit{IoU}                  & 0.479                          & 0.572                          & 0.541                               & \textbf{0.544} \\
                           \midrule
DUT-OMRON & \textit{Acc}                  & 0.878                          & 0.842                          & \textbf{0.883}     & 0.881                           \\
                           & \textit{IoU}                  & 0.451                          & 0.442                          & \textbf{0.509}     & 0.492                           \\
                           \midrule
DUTS      & \textit{Acc}                  & 0.890                          & 0.864                          & 0.893                               & \textbf{0.894} \\
                           & \textit{IoU}                  & 0.486                          & 0.478                          & \textbf{0.528}     & 0.524                           \\
                           \midrule
ECSSD     & \textit{Acc}                  & 0.905                          & 0.899                          & 0.915                               & \textbf{0.917} \\
                           & \textit{IoU}                  & 0.663                          & 0.672                          & \textbf{0.713}     & \textbf{0.713} \\ \bottomrule
\end{tabular}
\caption{A comparison of segmentation performance when different directions in the latent space are used to construct the training segmentation masks.}
\label{table:ablation_dual}
\end{table}
\begin{table*}[ht]
\small
\centering
\begin{tabular}{l|cccccccccc}
        \toprule
\multicolumn{1}{c}{} & \multicolumn{2}{c}{\textbf{CUB}} & \multicolumn{2}{c}{\textbf{Flowers}} & \multicolumn{2}{c}{\textbf{DUT-OMRON}} & \multicolumn{2}{c}{\textbf{DUTS}} & \multicolumn{2}{c}{\textbf{ECSSD}} \\ 
\multicolumn{1}{c}{}   &  Acc   &  IoU   &  Acc   &  IoU   &  Acc   &  IoU   &  Acc   &  IoU   &  Acc   &  IoU         \\ \midrule
$\lambda=0.1$ &  0.911 &  0.631 &  0.794 &  0.550 &  0.849 &  0.455 &  0.874 &  0.498 &  0.899 &  0.677 \\
$\lambda=0.2$ &  \textbf{0.919} &  \textbf{0.658} &  \textbf{0.782} &  \textbf{0.506} &  \textbf{0.880} &  \textbf{0.498} &  \textbf{0.891} &  \textbf{0.524} &  \textbf{0.912} &  \textbf{0.703} \\
$\lambda=0.4$ &  0.818 &  0.418 &  0.728 &  0.456 &  0.762 &  0.311 &  0.765 &  0.311 &  0.792 &  0.467 \\
$\lambda=0.8$ &  0.791 &  0.385 &  0.713 &  0.449 &  0.740 &  0.296 &  0.743 &  0.296 &  0.771 &  0.446 \\
\bottomrule
\end{tabular}
\caption{A comparison of segmentation performance for a BigBiGAN model when different values of $\lambda$ are used to find the latent vectors $v_l$ and $v_b$ in the optimization stage. 
Higher values of $\lambda$ yield latent directions $v$ that produce shifted images with greater variance in brightness between the center and outside pixels. 
Conversely, lower values of $\lambda$ yield latent directions $v$ that produce shifted images that align better to the original images.}
\label{table:ablation_lambda}
\end{table*}
\begin{figure}
\centering
  \includegraphics[width=0.47\textwidth]{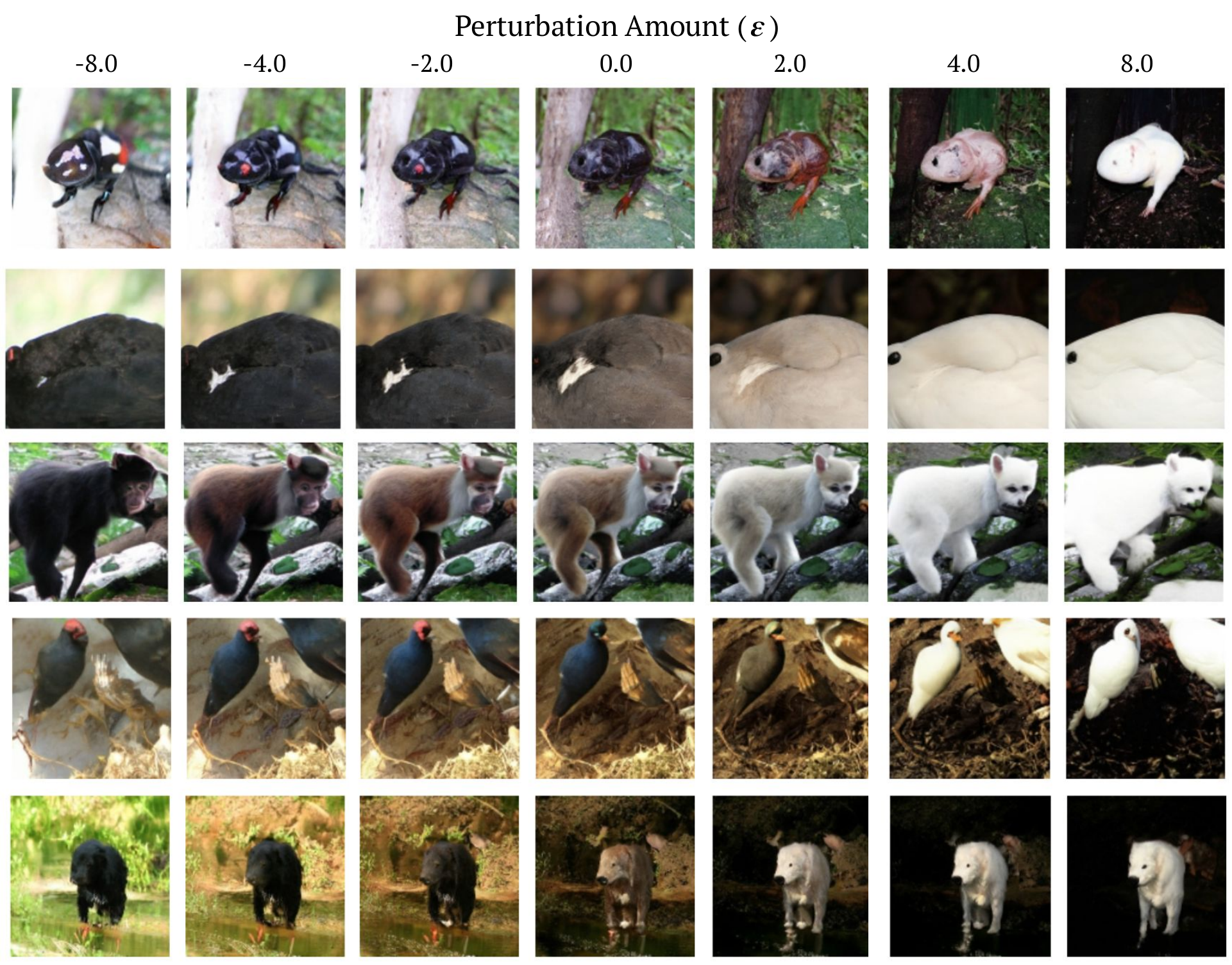}
  \caption{A comparison of generated images for different values of the perturbation length $\epsilon$, using the BigBiGAN generator. A value of $\epsilon = 0.0$ corresponds to the original image, with a random Gaussian latent vector $z \sim \mathcal{N}(0,1) $.} 
  \label{fig:perturbation_comparison}
\end{figure}
\begin{figure}[ht!]
  \centering
  \includegraphics[width=0.47\textwidth]{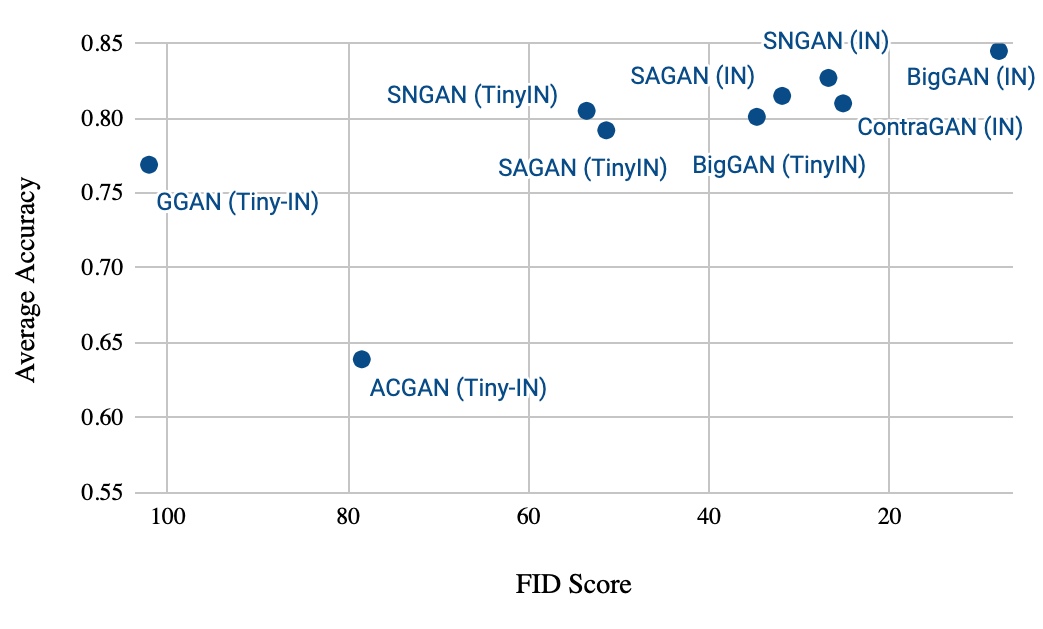}
  \caption{A plot of Frechet Inception Distance (FID) versus average segmentation accuracy across all five evaluation datasets (CUB, Flowers, DUT-OMRON, DUTS, ECSSD) for nine different GAN architectures. Lower FIDs are better (note that the x-axis is reversed). Lower FID scores correlate with improved final segmentation accuracy.}
  \label{fig:fid_comparison}
\end{figure} 
\begin{table}[ht]

\centering
\begin{tabular}{lr}
\toprule
\textit{GAN} & $v_s^T v_b$ \\ \midrule
BigBiGAN & -0.4376 \\
SelfCondGAN & -0.7854 \\
UncondGAN & -0.3522 \\
ContraGAN & -0.3297 \\
SAGAN & -0.4648 \\ \bottomrule
\end{tabular}
\caption{The dot product of the optimized foreground-lighter ($v_l$) and foreground-darker latent directions ($v_b$) for different generators, all of which have a $120$-dimensional latent space. Across the board, the directions are almost but not exactly antiparallel (random vectors in this space have an expected dot product of $0$ with variance $\frac{1}{120}$).}%
\label{table:how_antiparallel}
\end{table}

\section{Experiments}\label{s:experiments}

In this section, we present an extensive set of experiments demonstrating the method's performance, its wide applicability across image datasets, and its generalizability across deep generative architectures.

\subsection{Implementation details}
As our method is generator-agnostic, we apply our method to twelve generators, including three unconditional and nine conditional GANs.
For the three unconditional GANs (BigBiGAN~\cite{donahue19large}, SelfCondGAN~\cite{liu20diverse}, and UncondGAN~\cite{liu20diverse}), our procedure is completely unsupervised. For conditional GANs, our method is still unsupervised but the GAN naturally relies on class supervision for training.

To demonstrate the efficacy of our method across resolutions and datasets, we implement both, GANs trained on ImageNet~\cite{imagenet_dataset} at a resolution of 128px, and GANs trained on the smaller TinyImageNet dataset (100,000 images split into 200 classes) at a resolution of 64px.

All our experiments performed across all GANs utilize the same set of hyperparameters for both optimization and segmentation.
This is a key advantage of our method relative to other unsupervised/weakly-supervised image segmentation methods \cite{chen_unsupervised_2019,bielski_emergence_2019,benny_onegan_2020,arandjelovic2019object}, which are sensitive to dataset-specific hyperparameters.


All experimental details needed for reproducing our results are included in the the supplementary material and code will be made available.

\subsection{Evaluation data}

We evaluate the performance of our model on five datasets: DUT-Omrom~\cite{yang2013saliency}, DUTS~\cite{wang2017duts}, ECSSD~\cite{shi2016ecssd}, CUB~\cite{WelinderEtal2010}, and Flowers-102~\cite{nilsback09delving}.
The first three of these are standard benchmarks in the saliency detection literature and the remaining two are standard benchmark datasets in the object segmentation literature.

For purposes of comparison with prior methods, we evaluate on the saliency detection datasets using (pixel-wise) accuracy, mean intersection-over-union (IoU), and $F_\beta$-score with $\beta^2 = 0.3$.
Similarly, on the object segmentation datasets, we evaluate using accuracy, IoU, and $\text{max} F_\beta$.%
\footnote{$\text{max} F_\beta$ is the maximum $F_\beta$ score calculated using a range of binarization thresholds. We use 255 uniformly distributed binarization thresholds between $0$ and $1$, as in~\cite{voynov20big-gans}}.

\subsection{Results}

\paragraph{Performance on Benchmarks.}

In \cref{table:benchmark}, we compare our method to other recent work.
We emphasize that our method is uses the same model for all datasets and it has not seen any of the (training or evaluation) data for these datasets before.
Our method delivers strong performance across datasets, approaching and even outperforming some supervised/handcrafted saliency detection methods.

In comparison to~\cite{voynov20big-gans}, we perform better on four out of five benchmarks (all except CUB), even though we do not rely on humans to hand-pick latent directions.


In comparison to ReDo~\cite{chen_unsupervised_2019}, which trains separate GANs on CUB and Flowers-102, we perform worse on Flowers and better on CUB\@.
More generally, all of the approaches that involve training new layerwise GANs \cite{chen_unsupervised_2019,bielski_emergence_2019,benny_onegan_2020,arandjelovic2019object} are limited by their layerwise generator architectures and struggle to scale to complex datasets. For this reason, they are only trained on datasets consisting of images from a single domain with a single main subject, such as birds or flowers. It is unclear if it is even possible to successfully train a layerwise GAN on a diverse dataset such as ImageNet. By contrast, our ability to leverage pre-trained generators means that our method scales to complex and diverse datasets, such as those used for saliency detection.  

\paragraph{Performance across Generators.}

We investigate the generality of our method by performing the same optimization and training pipeline with twelve different GANs.
For each generator, we optimize to obtain a latent direction $v_l$, train a segmentation model using this direction, and evaluate its performance across the five datasets above.
The same hyperparameters are kept constant for \textit{all} GANs, including $\lambda = 0.2$ during the optimization phase.

Results are shown in \cref{table:gan_comparison}; BigBiGAN performs best, but all networks, even those using relatively weak TinyImageNet-trained GANs (e.g., GGAN~\cite{lim2017geometric}), deliver reasonable segmentation performance.
This highlights the benefits of our fully-automatic segmentation pipeline; our method performs well across a wide range of generators trained on different datasets at different resolutions.

Naturally, the quality of a final segmentation network produced by our method is related to the quality of the underlying generator.
\Cref{fig:fid_comparison} plots the Frechet Inception Distance (FID) score of nine conditional GANs versus the average accuracy of the corresponding segmentation networks produced by our method. Lower FID scores, which correspond to better GANs, correlate with improved accuracy.
This correlation suggests that as better GANs architectures are developed, our method will continue to produce better unsupervised segmentation networks.

Visual examples of perturbed images from different generators are shown in the supplementary material. 

\paragraph{Ablation: Comparing latent directions.}

We compare the performance of a segmentation masks using the two latent directions $v_b$ and $v_l$ together (\cref{eq:mask_gen_dual}), or each of them individually (\cref{eq:mask_gen_single}) visually in \cref{fig:perturbation_comparison}.
In \cref{table:ablation_dual}, we quantitatively compare the results of these three methods along with a fourth method in which we ensemble the final segmentation networks produced by $v_b$ and $v_l$ individually.

The foreground-lighter ($v_l$) and foreground-darker ($v_b$) directions yield similar results when used individually.
The combination ($v_l$ and $v_b$) provides superior results, on par with the ensemble. 
Unlike the ensemble, which requires training two networks, the combination of $v_b$ and $v_l$ adds minimal overhead compared to training with one direction.

We also compare actual latent directions $v_l$ and $v_b$. 
Due to the nonlinearity of the generator function, the optimal unit directions $v_l$ and $v_d$ are not necessarily negations of one another; indeed, we found in practice that they were close to but not exactly antiparallel. 
\Cref{table:how_antiparallel} gives exact numbers for a variety of different generator architectures.

\paragraph{Ablation: Varying $\lambda$ and $\epsilon$.}

The two hyperparameters in the optimization stage of our method are $\lambda$, which controls the trade-off between brightness and consistency, and $\epsilon$, which controls the magnitude of the perturbation. 
\Cref{table:ablation_lambda} compares segmentation results for BigBiGAN with $\lambda = 0.1, 0.2, 0.4, $ and $0.8$, with $0.2$ performing best.
We show an ablation for $\epsilon$ in the supplementary material. 

\vspace{-2mm}
\section{Conclusions}\label{s:conclusions}
We find that extracting a salient object segmentation from the latent space of a GAN is not only possible without supervision but also leads to state-of-the-art unsupervised segmentation performance on several benchmark datasets. 
In contrast to existing handcrafted features that have been engineered specifically for this task, we can extract this information from a network that has been trained for a very different purpose --- generating images. 
Surprisingly, we are able to generalize to a wide range of segmentation benchmarks without directly training on any real images. 
All training data has been generated by the GAN (and our method), suggesting that we can extract a generalizable and robust understanding of foreground and background.
Our findings prompt natural future research questions about what other concepts of the physical world can be automatically extracted from generative models, and to what extent we can use such extracted concepts to replace human supervision in other computer vision tasks.


\section*{Acknowledgments}
C.~R.~is supported by Innovate UK (project 71653) on behalf of UK Research and Innovation (UKRI) and by the European Research Council (ERC) IDIU-638009.
I.~L.~is supported by the EPSRC programme grant Seebibyte EP/M013774/1 and ERC starting grant IDIU-638009.
A.~V.~is funded by ERC grant IDIU-638009.

{\small\bibliographystyle{ieee_fullname}\bibliography{new,vedaldi_specific,vedaldi_general}}

\clearpage
\section*{Appendix}\label{s:appendix}

\subsection{Implementation Details}
   
\paragraph{Optimization.}

First, we optimize for the latent code vectors $v_d$ and $v_l$.
We generate latent codes $z \sim \mathcal{N}(0,1)$ and optimize the vector $v_l$ (or $v_d$) by gradient descent with the Adam~\cite{kingma14adam:} optimizer and learning rate $0.05$.
We use $\lambda = 0.2$ for the light direction $v_l$ and $\lambda = -0.2$ for the dark direction $v_b$.
We optimize perform $1000$ optimization steps, by which point $v_l$ (or $v_d$) has converged.

\paragraph{Segmentation.}

To generate data, we sample $z \sim \mathcal{N}(0,1)$, produce the mask $m = M(z)$, and refine the mask as described in the main paper.
We use a simple UNet~\cite{ronneberger15u-net:} with bilinear down/up-sampling as our segmentation network.
Following~\cite{voynov20big-gans}, we train for 12000 steps using Adam with learning rate $10^{-3}$ and batch size $95$, decaying the learning rate by a factor of $0.2$ at iteration $8000$.

\begin{figure*}
  \centering
  \includegraphics[width=0.5\textwidth]{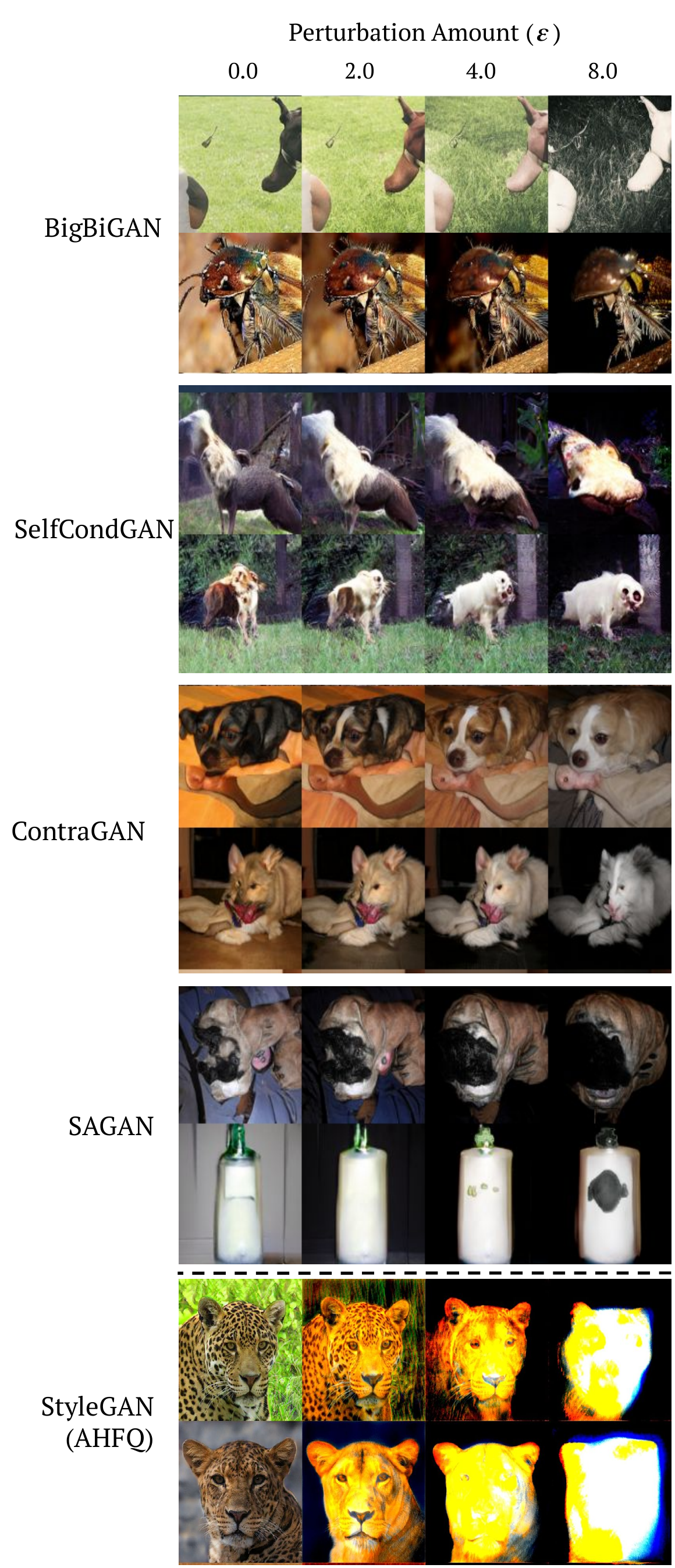}
  \caption{Examples of perturbed images generated by our method for five different generators (GANs). Note that the final generator, StyleGAN \cite{karras_style-based_2019}, is only trained on close-up portraits of animals, and thus cannot be used for general-perpose image segmentation. Nonetheless, our method successfully identifies the foreground and background of the generated animal portraits. } 
  \label{fig:gan_comparison}
\end{figure*}

\subsection{Datasets}

We apply a center crop to the object segmentation datasets, as in~\cite{voynov20big-gans}.

\begin{table}[H]
    \footnotesize
    \centering
    \begin{tabular}{lrrr}
        \toprule
        
        \textit{Dataset} & \textit{Num. Images} & \textit{Type} & \textit{Crop} \\
        \midrule
        CUB & 1000 & Object seg. & \cmark \\
        Flowers & 1020 & Object seg. & \cmark \\
        OMRON & 5168 & Saliency det. & \xmark \\
        DUTS & 5019 & Saliency det. & \xmark \\
        ECSSD & 1000 & Saliency det. & \xmark \\
        \bottomrule
    \end{tabular}
    \caption{Evaluation dataset statistics}%
    \label{table:eval_stats}
\end{table}

\subsection{Additional Ablations}

\begin{minipage}{2\linewidth}
    \centering
    \begin{table}[H]
        \small
        \centering
        \begin{tabular}{l|cccccccccc}
            \toprule
            \multicolumn{1}{c}{} & \multicolumn{2}{c}{\textbf{CUB}} & \multicolumn{2}{c}{\textbf{Flowers}} & \multicolumn{2}{c}{\textbf{DUT-OMRON}} & \multicolumn{2}{c}{\textbf{DUTS}} & \multicolumn{2}{c}{\textbf{ECSSD}} \\ 
            \multicolumn{1}{c}{}   &  Acc   &  IoU   &  Acc   &  IoU   &  Acc   &  IoU   &  Acc   &  IoU   &  Acc   &  IoU         \\ \midrule
            $\epsilon=1$ &  0.911          &  0.600          &  0.744          &  \textbf{0.600} &  0.867          &  \textbf{0.454} &  0.880          &  0.479          &  0.897          &  0.650 \\
            $\epsilon=2$ &  \textbf{0.912} &  \textbf{0.601} &  \textbf{0.773} &  0.479          &  \textbf{0.878} &  0.451          &  \textbf{0.890} &  \textbf{0.486} &  \textbf{0.905} &  \textbf{0.663} \\
            $\epsilon=4$ &  0.843          &  0.435          &  0.617          &  0.435          &  0.763          &  0.290          &  0.775          &  0.297          &  0.779          &  0.419 \\
            $\epsilon=6$ &  0.761          &  0.347          &  0.602          &  0.347          &  0.714          &  0.236          &  0.709          &  0.238          &  0.724          &  0.349 \\
            \bottomrule
        \end{tabular}
        \caption{A comparison of segmentation performance for a BigBiGAN-based model when different values of $\epsilon$ are used to find the latent vector $v_l$ in the optimization stage. Higher values of $\epsilon$ correspond to a greater-magnitude shift in the latent space during optimization.}
        \label{table:ablation_epsilon}
    \end{table}
\end{minipage}

\vspace{8mm}
In \autoref{table:ablation_epsilon}, we show ablation results for changing $\epsilon$ during the optimization process. Note that since the GAN used in this set of experiments (BigBiGAN) has a 120-dimensional latent space, the distribution of the norm of the $\mathcal{N}(0,1)$ latent vectors used to train the GAN is concentrated around (approximately) 11. That is to say, a shift of magnitude $\epsilon=6$ in the latent space is very large. 

\subsection{Additional Examples}

\paragraph{Across Datasets.}
In Figures 7-9, we show the results of applying our final segmentation network to random images from each of the five datasets on which we evaluated. 

\paragraph{Examples Across Generators.}
In Figure 10, we show examples of pairs of generated images and the corresponding extracted samples for a range of different GANs. 

\begin{figure*}
    \centering
    \includegraphics[width=0.8\textwidth]{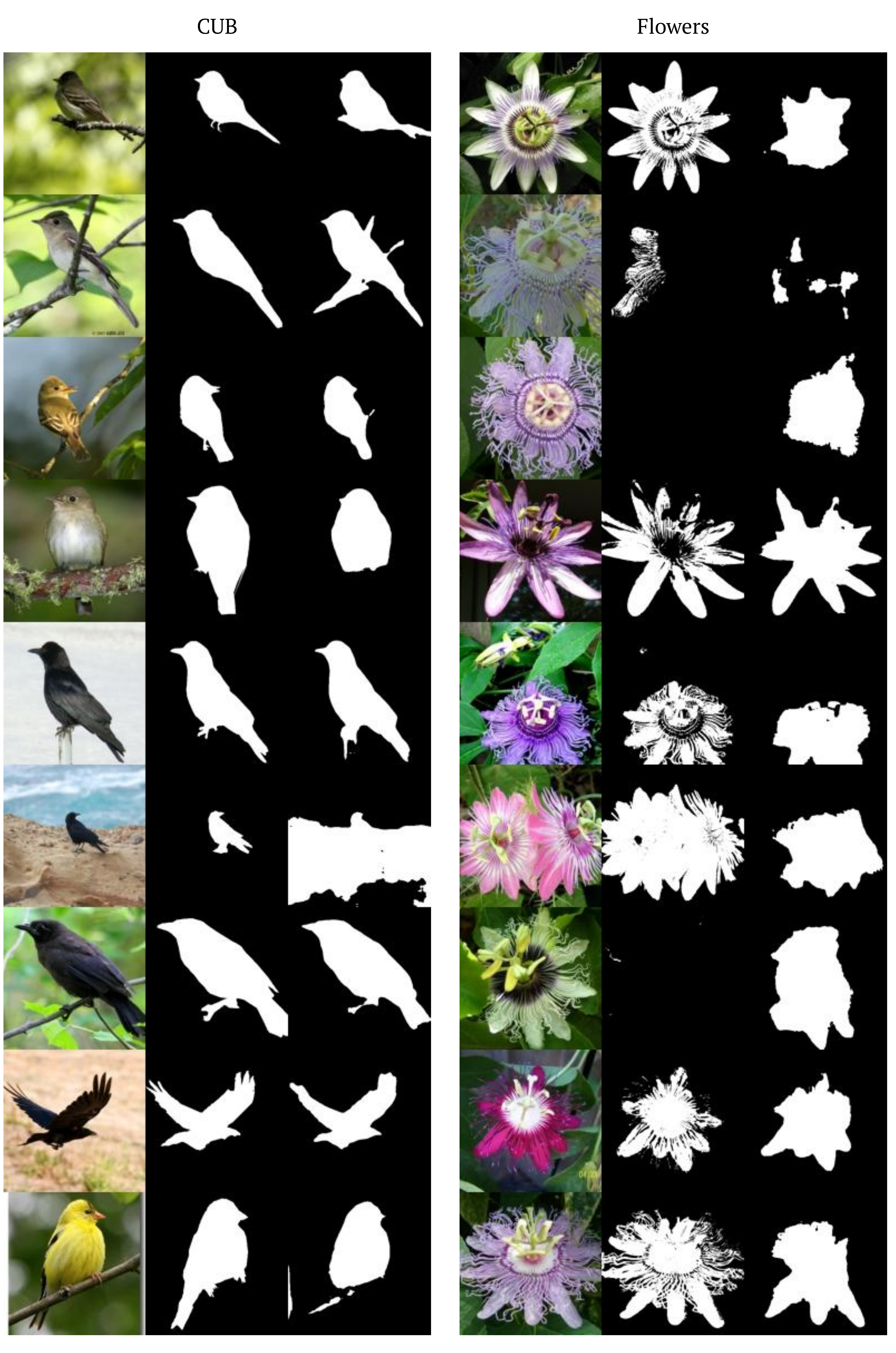}
    \caption{Examples of the final segmentation network across evaluation datasets. From left to right: original image, ground truth, prediction.}\label{fig:additional_examples_gans_1}
\end{figure*}
\begin{figure*}
    \centering
    \includegraphics[width=0.8\textwidth]{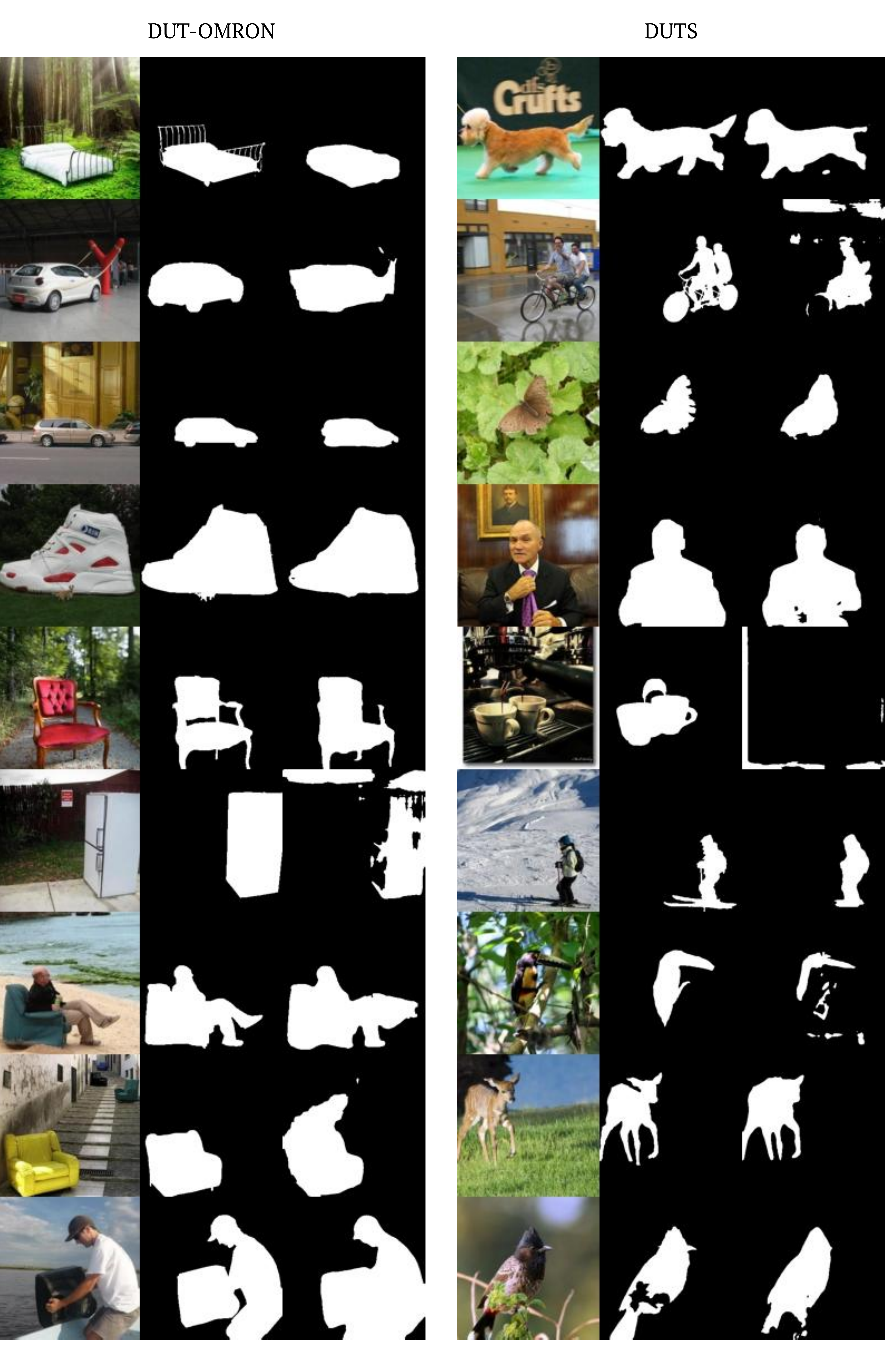}
    \caption{Examples of the final segmentation network across evaluation datasets. From left to right: original image, ground truth, prediction.}\label{fig:additional_examples_gans_2}
\end{figure*}
\begin{figure*}
    \centering
    \includegraphics[width=0.8\textwidth]{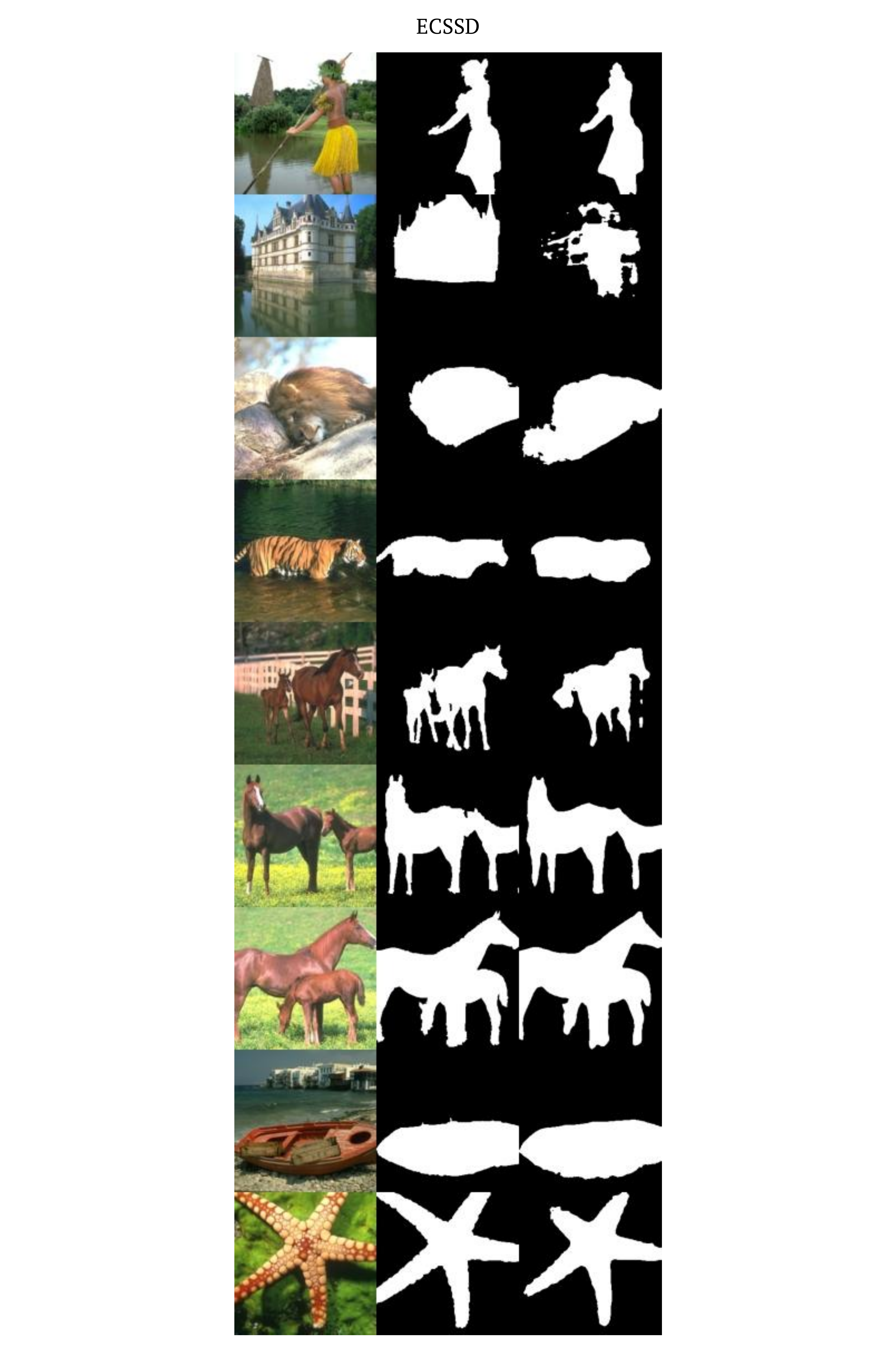}
    \caption{Examples of the final segmentation network across evaluation datasets. From left to right: original image, ground truth, prediction.}\label{fig:additional_examples_gans_3}
\end{figure*}
\begin{figure*}
    \centering
    \includegraphics[width=0.99\textwidth]{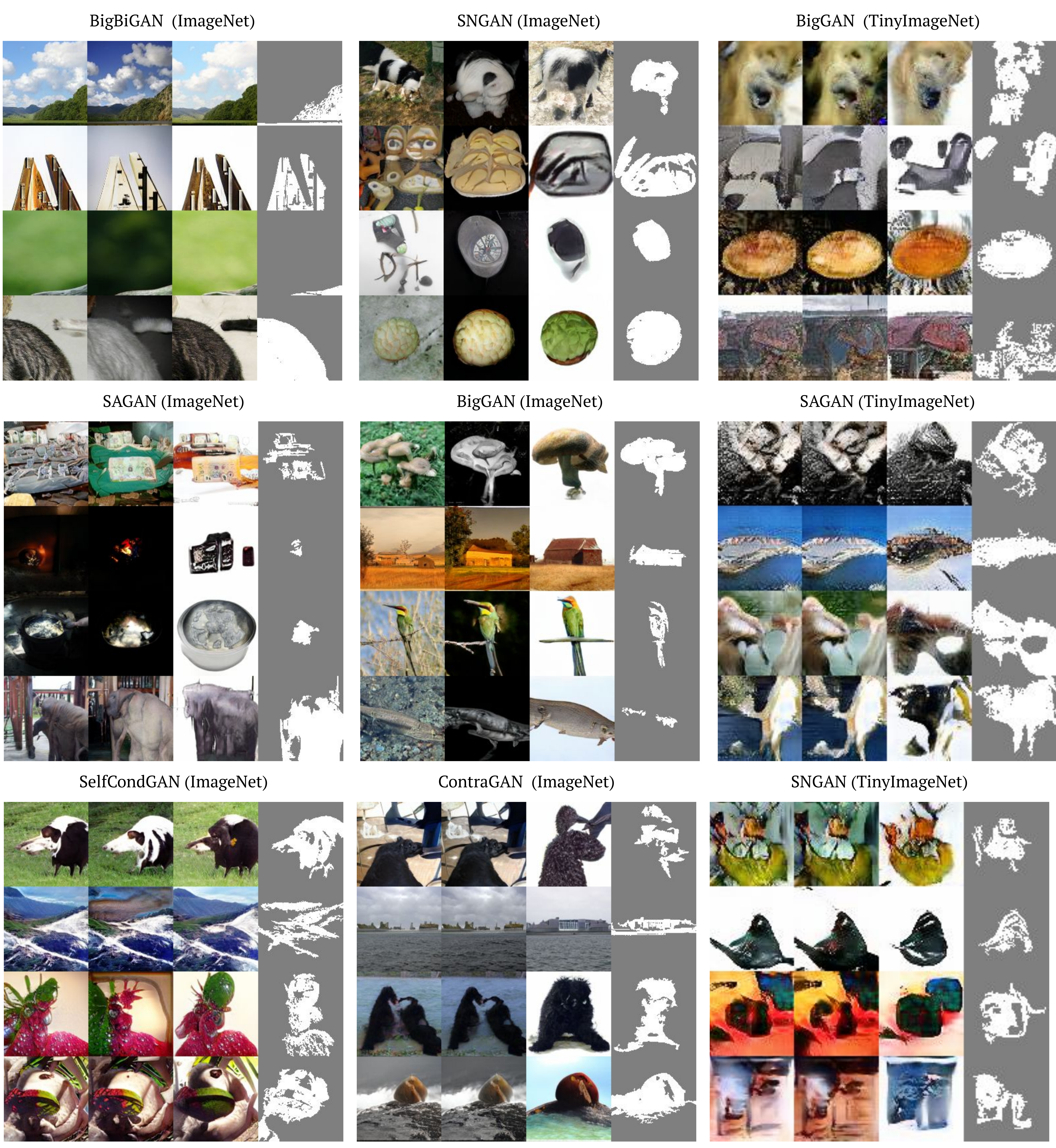}
    \caption{A comparison of perturbed images and their corresponding masks for many different generators.}\label{fig:additional_examples_datasets}
\end{figure*}

\end{document}